\documentclass[10pt,twocolumn,letterpaper]{article}

\usepackage{cvpr}
\usepackage{times}
\usepackage{epsfig}
\usepackage{graphicx}
\usepackage{amsmath}
\usepackage{amssymb}
\usepackage{caption}

\usepackage[breaklinks=true,bookmarks=false]{hyperref}

\cvprfinalcopy 

\ifcvprfinal\fi  
\begin{document}


\newcommand{\OURS}{RegNet}
\newcommand{\fang}[1]{{\color{red}[Fang: #1]}}
\newcommand{\han}[1]{{\color{blue}[Han: #1]}}
\newcommand{\MATTHIAS}[1]{{\color{red}\textbf{[Matthias: #1]}}}

\title{\OURS: Learning the Optimization of Direct Image-to-Image Pose Registration}

\author{Lei Han$^{1,2,3}$ \qquad Mengqi Ji$^{1,2}$ \qquad Lu Fang$^{1}$ \qquad Matthias Nie{\ss}ner$^{3}$ \vspace{0.2cm} \\
$^{1}$Tsinghua-Berkeley Shenzhen Institute \vspace{0.1cm}\\ $^{2}$Hong Kong University of Science and Technology \vspace{0.1cm}\\
$^{3}$Technical University of Munich 
}

\twocolumn[{%
	\renewcommand\twocolumn[1][]{#1}%
	\maketitle
	\begin{center}
		\includegraphics[width=\linewidth]{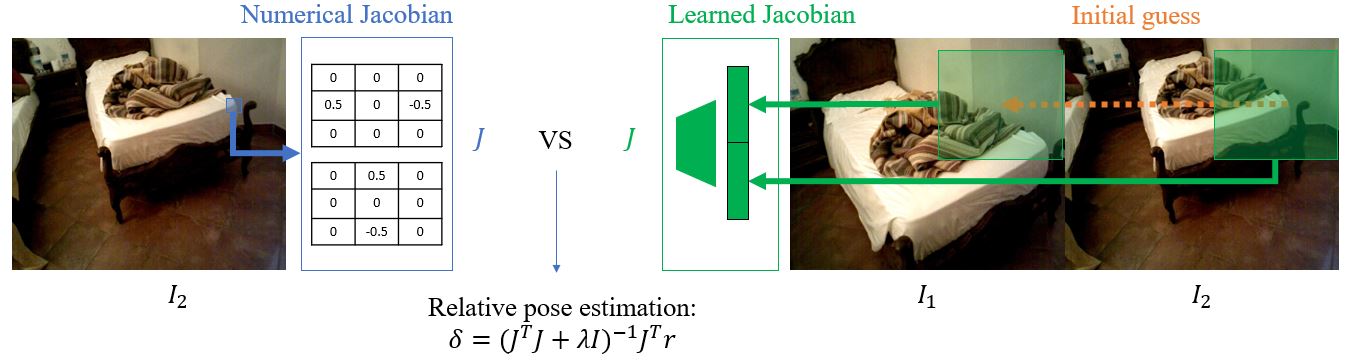}
		\captionof{figure}{Levenberg-Marquardt/Gauss-Newton optimization methods are widely-used for direct image alignment problems, where the Jacobian matrix is critical for fast and robust optimization. However, when initial alignment is not good enough, numerical Jacobian is unable to provide good gradient directions for the optimization procedure as only the neighboring pixels are employed in the computation of partial derivatives. To this end, we propose the learning of Jacobian based on the feature map of both images (which leverages a large receptive field as shown in green shadow). The joint learning of feature representation and Jacobian matrix enables us to have constraints on both the function and the partial derivatives in the optimization process, making it converges in fewer iterations and achieve more robust results.}
    	\label{fig:teaser}
	\end{center}
}]

\begin{abstract}

Direct image-to-image alignment that relies on the optimization of photometric error metrics suffers from limited convergence range and sensitivity to lighting conditions.
Deep learning approaches has been applied to address this problem by learning better feature representations using convolutional neural networks, yet still require a good initialization. In this paper, we demonstrate that the inaccurate numerical Jacobian limits the convergence range which could be improved greatly using learned approaches.
Based on this observation, we propose a novel end-to-end network, RegNet, to learn the optimization of image-to-image pose registration. By jointly learning feature representation for each pixel and partial derivatives that replace handcrafted ones (e.g., numerical differentiation) in the optimization step, the neural network facilitates end-to-end optimization. The energy landscape is constrained on both the feature representation and the learned Jacobian, hence providing more flexibility for the optimization as a consequence leads to more robust and faster convergence. In a series of experiments, including a broad ablation study, we demonstrate that RegNet is able to converge for large-baseline image pairs with fewer iterations. 

\end{abstract}

\section{Introduction}
Image-to-image relative pose estimation, a fundamental problem for many large projects such as structure from motion~\cite{schoenberger2016sfm}, visual SLAM applications~\cite{steinbrucker2011real}, etc, has been well studied in the past several decades. Usually, such a problem can be solved either relying on sparse, handcrafted features for indirect image registration~\cite{nister2004efficient} or direct methods that minimize the re-projection photometric error~\cite{alismail2016photometric,engel2013semi,baker2004lucas}. While feature-based methods suffer in texture-less or texture-repetitive environments, direct methods usually require good initialization, e.g., from features, to avoid local minima.

Data-driven schemes, based on recently popularized deep learning techniques, have shown great potential in assuring more robust performance~\cite{czarnowski2017semantic,ummenhofer2017demon}. As a result, recent works aim to estimate camera poses by either directly regressing a 6d vector for rotation and translation angels from convolutional neural network on image pairs~\cite{kendall2015posenet, ummenhofer2017demon, tateno2017cnn,melekhov2017relative} or leverage data-driven approaches to solve the pose estimation problem via minimizing a learned cost function~\cite{chang2017clkn, tang2018ba, goforth2018aligning}. Typically, regression-based methods cannot provide a confidence for the estimation results as the results are computed implicitly using neural networks, and the fixed network structure limits their possibility to be extended to multiple view settings. Optimization-based methods essentially learn feature representations for each pixel using neural networks and optimize the 'feature-metric' error instead of photometric error. Here, pixels are represented using RGB or gray-scale representations as conventional works~\cite{steinbrucker2011real,alismail2016photometric} and error is defined between the template image and warped target image. Although such methods can be easily extended to optimize the relative pose of multiple images and the feature representation may be improved significantly by adopting learned feature~\cite{chang2017clkn,tang2018ba}, the conventional non-linear Gauss-Newton method that serves to solve the optimization problem may strongly favor the smooth feature, leading to the over-smoothed feature along object boundaries.

In this paper, we exploit the power of data-driven approaches to jointly address the inappropriate feature representation and the sensitivity of initialization issues in the optimization step of direct image-to-image registration problems, where the poses are estimated by minimizing the feature-metric error. More specifically, by reviewing the optimization steps, we find that Jacobian matrix estimation using conventional methods (e.g., handcrafted Jacobian) tends to be inaccurate when the initial point is not good enough. We thus propose to learn the Jacobian matrix directly with a new neural network architecture under the supervision of re-projection error. The learned Jacobian scheme provides more flexibility and better weightings for each Jacobian entry, since the neural network learns optimization steps in an end-to-end fashion. 
Moreover, learning the Jacobian provides the possibility to reshape the original cost function and find a easier optimizable energy landscape, as we introduce constraints to both features and Jacobians. As a result, we achieve more robust and faster optimization convergence.
    
In order to verify the proposed algorithm, we present an end-to-end system for image-to-image pose registration problems, denoted as RegNet that jointly learns the feature representation and Jacobian for direct image-to-image registration. The joint learning of Jacobian and learned feature representation enables the neural network know the overall optimization step and can be trained end-to-end. Experiments on challenging dataset as well as ablation study demonstrate that the proposed RegNet with the novel learned Jacobian scheme achieves better performance in terms of accuracy with larger convergence range.

		\begin{figure*}[htbp]
		\begin{minipage}[b]{1.0\linewidth}
			\centering
			\includegraphics[width=18cm]{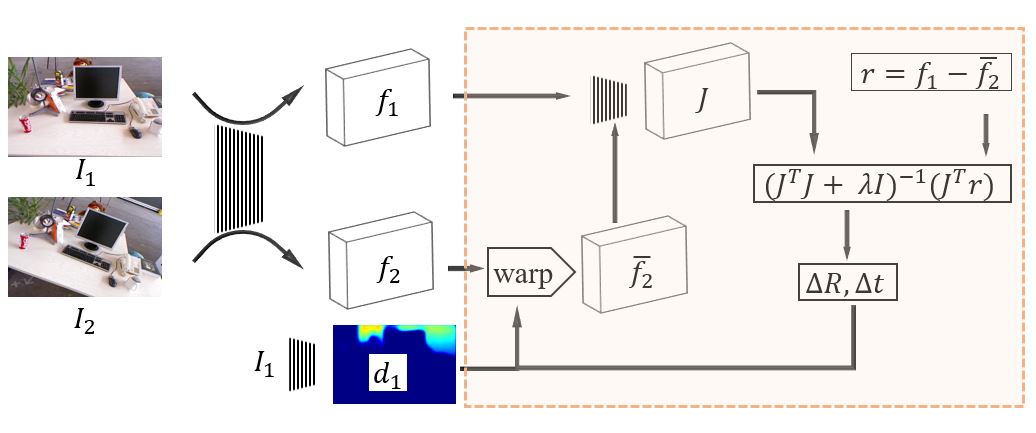}
		\end{minipage}
		\caption{Overview of RegNet. Given two input images $I_1$ and $I_2$, a depth prediction neural network~\cite{laina2016deeper} provides an initial depth estimation, which is then jointly optimized with the relative transformation iteratively using Levenberg-Marquardt optimization algorithm. Both feature and Jacobian matrix are learned using our new neural network to achieve robust image alignment.}
		\label{fig:SystemPipline}
	\end{figure*}

\section{Related Work}
Existing methods on image-to-image pose estimation problems can be roughly classified as three categories, namely feature-based, regression-based, and direct image alignment-based approaches.

Feature-based methods employ either handcrafted features (e.g., SIFT~\cite{lowe2004distinctive}) or learned feature (e.g., LIFT~\cite{yi2016lift}) to find correspondences between a provided image pair and make use of the essential matrix constraint to estimate the relative transformation matrix using robust optimization techniques like RANSAC~\cite{fischler1981ransac,brachmann2017dsac} or predict fundamental matrix directly~\cite{ranftl2018deep} given feature correspondences. Feature-based methods are robust to initialization and efficient for computing. However, such methods rely on feature detection and cannot make full use of information provided by the image, thus tend to overfit to feature-rich areas.

Regression-based methods, on the other hand, directly predict the rotation angle and normalized translation vector between the provided image pair using convolutional neural networks~\cite{melekhov2017relative, poursaeed2018deep} or the homography matrix given the provided image pair~\cite{detone2016deep}. For better generalization ability, Demon \cite{ummenhofer2017demon} takes as input optical flow vectors of each pixel and regresses both motion vectors and a dense depth map. Demon also provides a challenging dataset that contains large baseline image pairs including indoor and outdoor dataset, which is also adopted in this paper for fair comparisons. 
    
Given an input image pair $I_1$ and $I_2$, direct image alignment approaches aim to minimize the photometric error: $c = \sum_{x}||I_1(\tau(x,T,d)) - I_2(x)||^2$, where $x$ represents the pixel coordinate in image plane and $\tau(x,T,d)$ indicates the warped coordinate based on relative transformation $T$ between $I_1$ and $I_2$ as well as the depth value $d$ at pixel $x$ for projective transformation. DTAM~\cite{newcombe2011dtam} and DSO~\cite{engel2018direct} have demonstrated attractive results of joint estimating pose and depth by minimizing the photometric error using monocular cameras. However, they require specific motions for initialization and are sensitive to lighting changes / fast motion due to the optical flow assumption made in their optimization step, which assumes that the photometric term at the same place are constant and continuous. Deep learning approaches are adopted to overcome these problems; however, they are still at an early stage.
CLKN~\cite{chang2017clkn} adopt the Lucas-Kanade algorithm~\cite{baker2004lucas} to minimize the feature-metric error of corresponding pixels for estimating the 2D transformation of two image patches, where features are learned using pyramid convolutional neural networks and the Gauss-Newton optimization step in each iteration is regarded as ``LK'' layer which could be used for back propagation. \cite{goforth2018aligning} apply similar method for aligning satellite images captured at different seasons. For relative pose estimation in 3D space, depth information is also required for image warping, which is unknown in most cases and needs to be estimated jointly with motion parameters. Both CodeSLAM~\cite{bloesch2018codeslam} and BA-Net~\cite{tang2018ba} are proposed to use a more compact representation for depth maps instead of maintaining depth values for each pixel independently. CodeSLAM~\cite{bloesch2018codeslam} proposes a compact code to represent the depth map using variational auto-encoder networks and optimize the photometric error on the code and motion parameters. BA-Net~\cite{tang2018ba} makes use of a depth prediction network~\cite{laina2016deeper} to infer the depth map as a combination of depth basis and jointly estimates the weight of each depth basis as well as motion by minimizing the feature metric error learned from~\cite{yu2017dilated} for two-view bundle adjustment problem. \cite{clark2018ls} adopts a regression network for motion and depth map initialization, and solves the nonlinear least square problem of jointly estimating depth per pixel and motion using a recurrent neural network. 
    
In this work, we investigate the image-to-image relative pose estimation problem using direct methods as shown in Fig.~\ref{fig:SystemPipline}. For monocular image alignment problems, depth input is required for image warping which is estimated using a depth prediction neural network~\cite{laina2016deeper} like~\cite{tang2018ba,tateno2017cnn}. To further account for the inaccurate depth prediction from monocular images, we also refine the predicted depth map in a joint depth and motion optimization stage as~\cite{tang2018ba}. While direct optimization based methods suffer from limited convergence range, our learned Jacobian replaces the handcrafted numerical Jacobian, and thus achieves more robust and accurate alignments.

\section{RegNet}
\subsection{Problem Formulation}
\label{sec_PF}
Given an image pair ($I_1$, $I_2$), we aim to estimate their relative transformation ${T}$ in SE3 space. Depth of each pixel is required for image warping under 3D transformations, which is predicted using a monocular depth prediction network~\cite{laina2016deeper}. The predicted depth map is inaccurate and may introduce bias for image alignment. Instead of optimizing the depth of each pixel, we adopt the compact depth map representation method of BANet~\cite{tang2018ba}, which predicts $N$ depth basis $\mathbf{B}$ and represent the depth map $D$ using a weight vector $\mathbf{w}$: $D = ReLU(\mathbf{w}^T\mathbf{B})$. $\mathbf{w}$ can be jointly optimized together with motion $T$ by minimizing the feature-metric error $\mathbf{r}$:
\begin{equation}
    \mathbf{r} = [r_0, r_1, \dots, r_{M-1}],
    \label{Eqn_feature_metric}
\end{equation}
where residual $r_i = f_1(x_i) - f_2(\tau(x_i,T,D))$ indicates the feature-metric error of pixel $x_i$, and $M$ is the number of pixels determined by image resolution. $x_i$ represents the pixel coordinate at image $I_1$, and is warped to $I_2$ by function $\tau(x_i,T,D)$. $f_k = F(I_k), k = 0, 1$ indicates the learned feature representation from image $I_k$ using the feature learning network proposed in Sec.~\ref{sec_FLN}. 

Given Eqn. (\ref{Eqn_feature_metric}), $\{ \mathbf{w}, {T}\}$ are estimated by minimizing the L2 norm of feature-metric error:
\begin{equation}
\label{eqn_cost}
   \{ \mathbf{w}^*, {T}^*\} = \arg \min_{\mathbf{w},{T}}  ||\mathbf{r}||^2,
\end{equation}
which can be optimized using LM algorithm~\cite{gill1978algorithms}:
\begin{align}
\label{eqn_opt}
    \mathbf{\delta} &= (\mathbf{J}^T\mathbf{J} + \lambda \mathbf{I})^{-1}(\mathbf{J}^T\mathbf{r}),
\end{align}
where 
\begin{equation}
\begin{aligned}
    \mathbf{J} &= \frac{\partial \mathbf{r}}{\partial(\mathbf{w},T)}\\
    &= [\mathbf{J_w} \quad \mathbf{J_T}]^T \\
               &= [\frac{\partial\mathbf{r}}{\partial\mathbf{w}} \quad \frac{\partial\mathbf{r}}{\partial{T}}]^T.
\end{aligned}
\label{Eqn_J}
\end{equation}

Recall that $\mathbf{r} = [r_0, r_1, \dots, r_{M-1}]$. For ease of presentation, we will only consider the Jacobian matrix $J_k$ with respect to $r_k$ in the following discussions. To further simplify notations, we denote $y_i = (u_i,v_i) = \tau(x_i,T,D)$ as the warped coordinate of $x_i$. Then, the $\frac{\partial{r_i}}{\partial{T}}$ and $\frac{\partial{r_i}}{\partial{\mathbf{w}}}$ in Eqn. (\ref{Eqn_J}) can be represented as follows:
\begin{equation}
\label{eqn_drdt}
\begin{aligned}
    \frac{\partial{r_i}}{\partial{T}} &=
    -\frac{\partial{f_2(y_i)}}{\partial{T}} \\
    &=-[\frac{\partial{f_2(y_i)}}{\partial{u_i}} \quad \frac{\partial{f_2(y_i)}}{\partial{v_i}}]
    [\frac{\partial{u_i}}{\partial{T}} \quad \frac{\partial{v_i}}{\partial{T}}]^T 
\end{aligned}
\end{equation}
\begin{equation}
\label{eqn_drdw}
\begin{aligned}
\frac{\partial{r_i}}{\partial{\mathbf{w}}} &=
    -\frac{\partial{f_2(y_i)}}{\partial{\mathbf{w}}} \\
    &= -\frac{\partial{f_2(y_i)}}{\partial{D}} 
    \frac{\partial{D}}{\partial{\mathbf{w}}}\\
    &= -[\frac{\partial{f_2(y_i)}}{\partial{u_i}} \frac{\partial{f_2(y_i)}}{\partial{v_i}}]
    [\frac{\partial{u_i}}{\partial{D}} \frac{\partial{v_i}}{\partial{D}}]^T 
    \frac{\partial{D}}{\partial{\mathbf{w}}}.
\end{aligned}
\end{equation}
While $\frac{\partial{u}}{\partial{T}}$, $\frac{\partial{v}}{\partial{T}}$, $\frac{\partial{u}}{\partial{D}}$, $\frac{\partial{v}}{\partial{D}}$ can be computed analytically in SE3 space~\cite{blanco2010tutorial}, $\frac{\partial{f_2}}{\partial{u}}, \frac{\partial{f_2}}{\partial{v}}$ need to be estimated using numerical differentiation from neighboring pixels:
\begin{equation}
\label{eqn_feature_grad}
\begin{aligned}
    \frac{\partial{f}}{\partial{u}} &= \frac{f(u+1,v) - f(u-1,v)}{2},  \\
    \frac{\partial{f}}{\partial{v}} &= \frac{f(u,v+1) - f(u,v-1)}{2}.
\end{aligned}
\end{equation}

Re-examining the previous formulations, when precise initialization is provided, the Jacobian based on the handcrafted formulation (denoted as \textit{Numerical Jacobian}) is accurate enough to further refine the cost function. However, when initial estimation is far away from the ground truth, the fixed numerical Jacobian lacks the ability to account for the linearization of the cost function in Eqn.~(\ref{eqn_cost}). To address this shortcoming, we propose RegNet to learn the entire Jacobian matrix $\mathbf{J}$ with a new deep neural network based on the feature representations of the image pair. As demonstrated by the ablation study and experiments on public datasets in Sec.~4, our proposed RegNet solves the non-linear iterative optimization more effectively. In the following subsections, we elaborate the Feature Learning Network (FLN) and the Jacobian Prediction Network (JPN), respectively. 
	\begin{figure*}[htbp]
		\begin{minipage}[b]{1.0\linewidth}
			\centering
			\includegraphics[width=18cm]{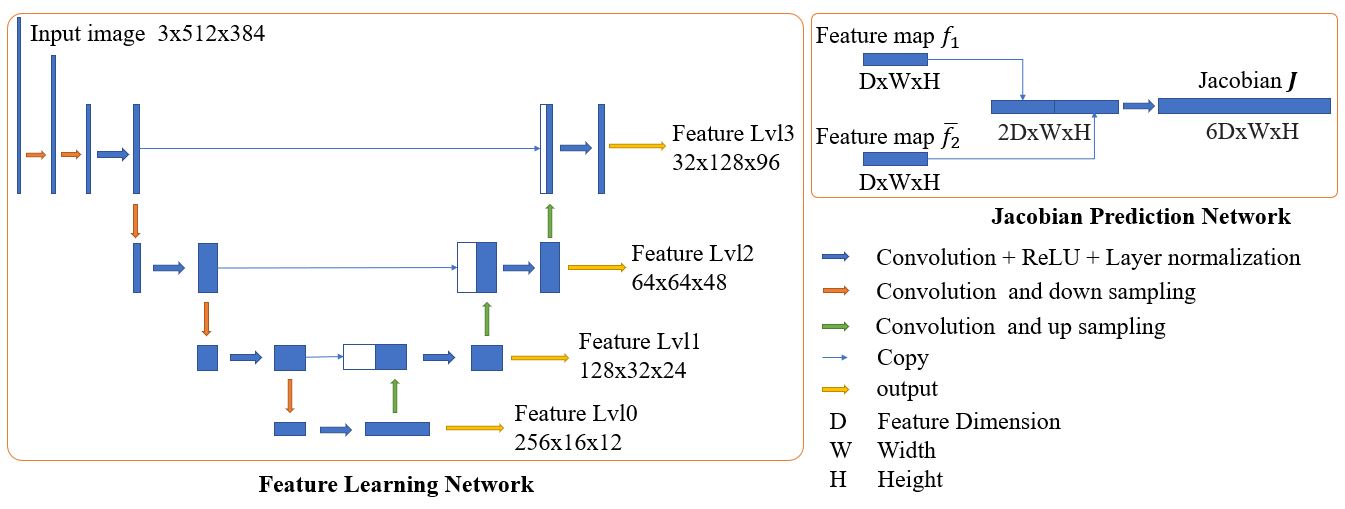}
		\end{minipage}
		\caption{Details of the proposed network are presented in this figure. For feature learning network, we use a U-Net like structure to predict feature maps at different resolutions for coarse-to-fine image alignment. Image pair are resized to 512x384 firstly as input. For down sampling we use average pooling after Gaussian smoothing and up sampling with bilinear interpolation. Jacobian prediction network takes input with both the feature map of the first image $f_1$ and warped feature map of the second image $\bar{f_2}$ and predicts the entire Jacobian matrix.}   
		\label{fig:FeatureNet}
	\end{figure*}
	
\subsection{Feature Learning Network}
\label{sec_FLN}
Direct image alignment using photometric error has been well studied in previous approaches~\cite{alismail2016photometric,engel2018direct}, which however are not robust to luminance change due to the photometric consistency assumption. Learning feature representations to replace the photometric term are then proposed in \cite{czarnowski2017semantic,chang2017clkn,tang2018ba} to improve the robustness for challenging cases. Similarly, we leverage the powerful feature representations to replace photometric error term using feature-metric error. To this end, we propose a multi-layer feature extraction network with U-net ~\cite{ronneberger2015unet} connections as shown in Fig.~\ref{fig:FeatureNet}. The coarsest level $l_0$ layer is down sampled using average pooling layers after convolution to a resolution of $16 \times 12$ with 256 dimensions for initial alignment. Feature maps are then up sampled using bilinear interpolation by a scale of 2 and expanded with skip connections for each level, namely $l_1$ feature map at the resolution of $32 \times 24$ with 128 dimensions, $l_2$ feature map at the resolution of $64 \times 48$ with 64 dimensions, and finally $192 \times 128$ with 32 dimensions. In our experiments, we use feature maps from $l_0$ to $l_2$ for coarse registration without depth optimization. Based on the initial pose provided from the $l_2$ layer, feature map $l_3$ is used for precise registration along with depth map optimization.

\subsection{Jacobian Prediction Network}
\label{sec_JPN}

Reviewing the formulation and optimization elaborated in Sec.~\ref{sec_PF}, the Jacobian matrix in Eqn.~(\ref{eqn_opt}) is typically computed based on the numerical differentiation in Eqn.~(\ref{eqn_feature_grad}) \cite{tang2018ba,tateno2017cnn}. 
Regardless the robustness of learned features in handling brightness and view angle as noted in~\cite{czarnowski2017semantic}, the direct optimization framework remains suffering from limited convergence range and requires good initialization for accurate alignment. The explanation to this issue can be analyzed as follows.

Recall that Jacobian matrix $\mathbf{J}$ is widely used in the Gauss-Newton/LM optimization step to estimate the Hessian matrix $\mathbf{H} = 2\mathbf{J}^T\mathbf{J}$ and gradient vector $\mathbf{g} = 2\mathbf{J}^T\mathbf{r}$. 
The chain rule is adopted to compute $\mathbf{J}$ as shown in Eqn.~\ref{eqn_drdt}. 
While $\frac{\partial u}{\partial T}, \frac{\partial v}{\partial T}$ can be computed analytically on the Lie manifold, $\frac{\partial f}{\partial u}$ and  $\frac{\partial f}{\partial v}$ are approximated using numerical differentiation. 
However, such approximation holds under the case that feature is smooth between the initialization pixel and ground truth corresponding pixel. 
Learning features that are suitable for Gauss-Newton optimization enforce the learned feature map to be smooth across large baseline, which blurs the feature representation at object boundaries certainly.

Instead of using numerical differences of $f_2$ in the non-linear Gauss-Newton optimization, we learn the computation of the partial derivatives of the feature-metric error.
This leads to a better-shaped energy landscape compared to previous methods that only learn feature representations with handcrafted Jacobian (denoted as numerical Jacobian). The advantage originates from the learnable parameters for the respective partial derivatives, which remove the constraint of smoothness on feature representations effectively. 
We also argue that the Jacobian matrix $\mathbf{J}$ could be better computed based on $f_1$ and the warped feature map of $I_2$: $\bar{f_2}$, as the cost function is nonlinear yet the handcrafted Jacobian computation only considers the linearization step of $f_2$. 
In other words, introducing the information of $f_1$ in the neural network helps to provide more constraints (e.g., the similarity of $f_1$ and $\bar{f_2}$) during the computation of Jacobian matrix. As a consequence, we can `navigate' on the energy landscape more robustly (so as to avoid local minima), and the convergence can be achieved faster due to the learned Jacobian entries. 

As illustrated in Fig. \ref{fig:FeatureNet}, the Jacobian prediction network is realized by concatenating the feature map $f_1$ and warped feature map of $f_2$: $\bar{f_2}$ and then followed by four residual blocks from ResNet~\cite{he2016deep}. 
Note that the last layer does not have nonlinear activation and output both positive and negative Jacobian entries.


\subsection{Training details}
To estimate the image-to-image pose, we jointly optimize the relative transformation ${T}$ and image depth $D$ using direct optimization techniques~\cite{tang2018ba}. For depth prediction network, ~\cite{laina2016deeper} is adopted for depth basis prediction, and it is pretrained on the provided training set to provide rough depth map initialization. We then fine-tune these results as well as the feature learning network together for joint depth-motion optimization. 
The feature learning network is trained in a hierarchical manner. 
Firstly, we only predict the $l_0$ layer feature map and trained jointly with Jacobian prediction network for the non-linear Gauss-Newton optimization. 
Based on the initial relative transformation provided from previous layer, the following layer is trained to refine the results. 

For the Jacobian prediction network, it is trained on the first layer for initialization merely, while later layers are optimized using numerical Jacobian as they are already roughly aligned to convergence range. 
During training, instead of using Identity matrix as initial relative transformation matrix, we provide a random disturbance on it (e.g., randomly rotate each axis from -10 to 10 degrees and translate with a norm of 10\% of the average value of the predicted depth map) for each training image pair in order to avoid overfitting and enforce the network to be able to deal-with large motion cases. 
The network is trained in a supervised manner given ground truth of relative transformation $T^*$ and depth map $D^*$ by minimizing the re-projection error for each pixel:
\begin{equation}
\label{eqn_reprojection_error}
    L =\sum_{i=0}^{M-1} ||\tau(x_i,T,D) - \tau(x_i,T^*,D^*)||^2,
\end{equation}
where $T, D$ are the predicted relative transformation matrix and depth map, respectively. In practice, we find that when set the $l2$ norm of reprojection error as the cost function and train the network from random initialization directly, the network learns slowly and requires too many training samples to converge. A better solution is to replace the $l2$ norm $L$ by $log(L + 1)$, which gives negative samples (image pairs that have a large reprojection error after optimization) a smaller weight and encourages positive samples during training. After the bootstrapping process, we further replace the $log(L+1)$ back to $L$ to deal with large baseline image pairs. 

For the final layer where the depth and motion are jointly optimized, we combine the cost from image alignment (reprojection error) and depth map prediction to jointly train the depth prediction network and feature learning network:
\begin{equation}
    \label{eqn_full_error}
    C = \lambda L + \beta(D_1,D^*) + \beta(D_0, D^*),
\end{equation}
where $\beta(D,D^*)$ is the BerHu penalty between the dense depth map $D$ and $D^*$, $D_0$ indicates the initially estimated depth map, and $D_1$ is the optimized depth map. $\lambda$ is a fixed parameter to balance image alignment error and depth optimization error.

The network is trained using Adam optimizer~\cite{kingma2014adam} with an initial learning rate of $0.0001$ using pytorch~\cite{paszke2017automatic}. $80K$ image pairs are randomly sampled from the training set to train each layer of the feature learning network.

\section{Experiments}

The dataset provided by Demon~\cite{ummenhofer2017demon} is adopted for both training and testing. It contains randomly selected image pairs from \textit{MVS}~\cite{fuhrmann2014mve} for outdoor dataset, \textit{RGBD}~\cite{sturm2012benchmark} and \textit{SUN3D}~\cite{xiao2013sun3d} for indoor dataset. Note that the synthetic dataset \textit{Scenes11} with random geometry used in~\cite{ummenhofer2017demon} is excluded, as it is not applicable for the depth prediction networks~\cite{laina2016deeper}. 

According to Demon~\cite{ummenhofer2017demon}, the training data contains 3773 image pairs for \textit{RGBD} dataset, 1346 image pairs for \textit{MVS} dataset and 32885 image pairs for \textit{SUN3D} dataset. To make sure that the training data used for three dataset are basically equal, we use different sampling rates for different dataset, namely 1 for \textit{MVS} dataset, 0.5 for \textit{RGBD} dataset and $0.05$ for \textit{SUN3D}, respectively.

\subsection{Quantitative Evaluation}
We compare RegNet with Demon~\cite{ummenhofer2017demon}, the state-of-the-art image-to-image registration approach that regresses relative transformation and dense depth map based on optical flow vectors. 
Experiments using conventional methods implemented in Demon are also included; e.g., Base-SIFT which uses sparse SIFT feature points for correspondence search, 
Base-FF which uses the dense optical flow method~\cite{bailer2015flow}, Base-Matlab using the KLT tracker implemented in Matlab, and Base-Mat-F employing the optical flow fields estimated from Demon. 
The collected correspondences are adopted to estimate the essential matrix using RANSAC~\cite{fischler1981ransac} and 8-point algorithm~\cite{hartley1995defence}, which are further refined by minimizing re-projection error of correspondences. 

We use the rotation error and translation vector error adopted in~\cite{ummenhofer2017demon} for evaluations. 
The mean error of each algorithm is presented in Tab.~\ref{tab:meanError}. 
Compared with other algorithms, the proposed RegNet (denoted as Ours-LJ as it uses learned Jacobian scheme) achieves considerably better performance on the \textit{RGBD} and \textit{SUN3D} dataset, while comparable performance on \textit{MVS} dataset. 
The possible explanation to the limited improvement on \textit{MVS} dataset could be attributed to the depth prediction network~\cite{laina2016deeper} for single image depth prediction, which relies on the labeled ground truth depth during training. 
However, for the dataset of Demon, the infinity depth (such as the sky) is set as $NAN$, leading to unconstrained depth prediction results on the outdoor dataset \textit{MVS}. As further demonstrated by the median error of rotation and translation in Tab.~\ref{tab:medianError}, Ours-LJ results in significantly higher performance for \textit{MVS} dataset as well. 

For more comprehensive evaluation of the proposed Jacobian learning scheme, we train a network that learns feature using the same network structure as RegNet, while the Jacobian using numerical Jacobian scheme as~\cite{tang2018ba} instead of learning (denoted as Ours-NJ). 
For all the three dataset, the joint learning of feature and Jacobian (Ours-LJ) leads to considerably better results compared with methods that learn features merely (Ours-NJ). In other words, the proposed Jacobian prediction network takes effect to learn the optimization in a better way than numerical Jacobian.    

\begin{table}[htbp]		\newcommand{\tabincell}[2]{\begin{tabular}{@{}#1@{}}#2\end{tabular}}
\centering
\caption{Quantitative evaluations on public data-set~\cite{ummenhofer2017demon} (mean error in degree). Ours(NJ) indicates that \textbf{N}umerical \textbf{J}acobian is used, and ours(LJ) shows the result of \textbf{L}earned \textbf{J}acobian.}
	\begin{tabular}{|l|c|c|c|}
	\hline
	& \tabincell{c}{\textit{RGBD} \\ T / R} & \tabincell{c}{\textit{MVS} \\ T / R} & \tabincell{c}{\textit{SUN3D} \\ T / R}  \\
	\hline
	Base-SIFT & $56.02/ 12.01$ & $60.52/21.18$ & $41.83 / 7.70$\\	
	\hline
	Base-FF & $46.06 / 4.71$ & $17.25 / \textbf{4.83}$ & $33.30 / 3.68$\\	
	\hline
	Base-Matlab & $49.61 / 12.83$ & $32.74 / 10.84$ & $32.30 / 5.92$\\	
	\hline
	Base-MatF & $22.52 / 2.92$ & $18.54 / 5.54$ & $26.33 / 2.23$\\	
	\hline
	Demon & $20.59 / 2.64$ & $\textbf{14.45}/{5.15}$ & $18.81 / {1.80}$\\	
	\hline \hline
    Ours-NJ & ${15.6} / {2.67}$ & $23 / 9.0$ & ${19.0} / 2.47$\\	
    \hline
    Ours-LJ& $\textbf{11.25} / \textbf{2.16}$ & $ 16.23 / 5.95$ & $\textbf{14.77} / \textbf{1.75}$\\	
    \hline
	\end{tabular}
    \label{tab:meanError}
	\end{table}

	\begin{table}[htbp]		\newcommand{\tabincell}[2]{\begin{tabular}{@{}#1@{}}#2\end{tabular}}
	\centering
	\caption{Quantitative evaluations on Public data-set (median error in degree)}
	\begin{tabular}{|l|c|c|c|}
	\hline
	& \tabincell{c}{\textit{RGBD} \\ T / R} & \tabincell{c}{\textit{MVS} \\ T / R} & \tabincell{c}{\textit{SUN3D} \\ T / R}  \\
	\hline
	Demon & $10.33 / 1.51$ & $ 6.49/1.53$ & $\textbf{11.26} / 1.40$\\	
    \hline
    Ours-LJ & $\textbf{5.38} / \textbf{0.81}$ & $\textbf{5.30} / \textbf{1.35}$ & $11.68 / \textbf{1.25}$\\	
    \hline
	\end{tabular}
    \label{tab:medianError}
	\end{table}

\subsection{Ablation Study}
To further evaluate the performance of joint feature and Jacobian learning network, we particularly study the following four methods using the same test data:
\begin{itemize}
    \item \textit{Conventional} - only the photometric error is minimized using numerical Jacobian. 
    \item \textit{Learned Feature} - only features are learned while the feature metric error is minimized using numerical Jacobian scheme during both training and testing.
    \item \textit{RegNet} - the method proposed in this paper, where both Jacobian and feature space are learned.
    \item \textit{RegNet with Numerical Jacobian} - feature representations are learned jointly with the learned Jacobian scheme and trained as \textit{RegNet}, yet Numerical Jacobian is used instead during testing to verify the performance of the Jacobian prediction network.
\end{itemize}

	\begin{figure}[htbp]
		\begin{minipage}[b]{1.0\linewidth}
			\centering
			\includegraphics[width=8cm]{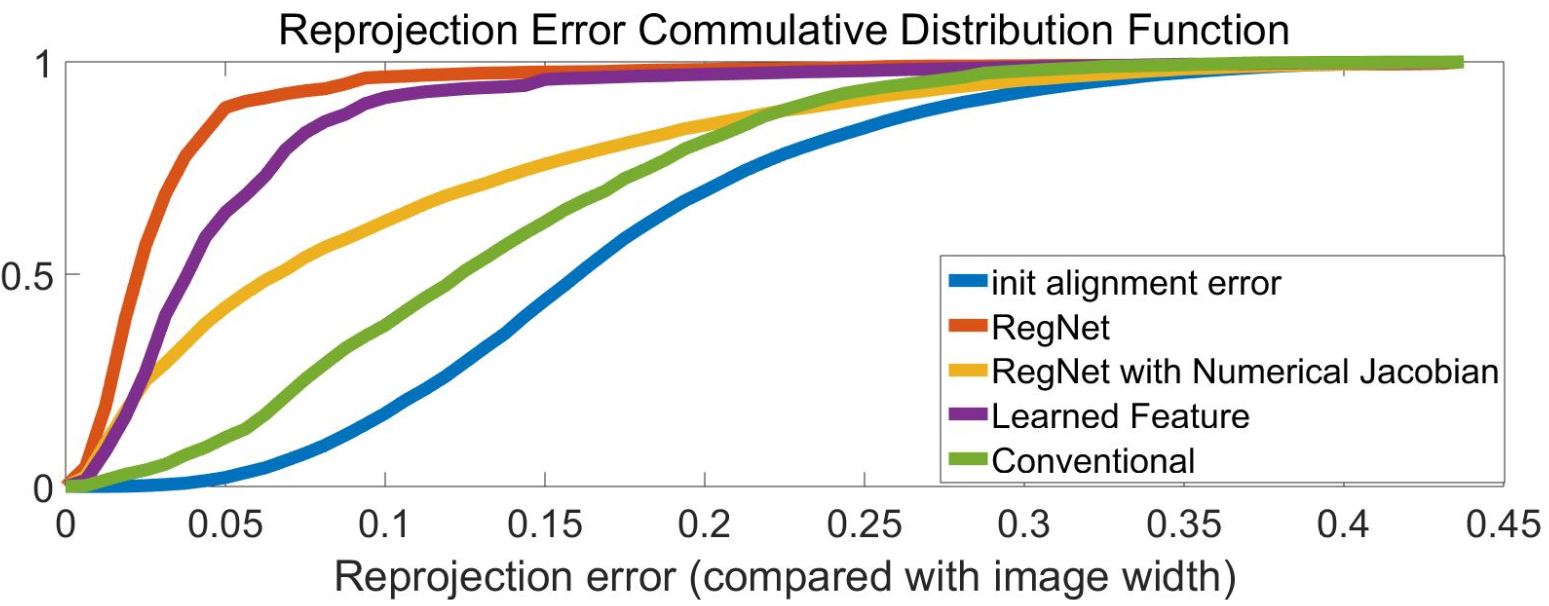}
		\end{minipage}
		\caption{Illustration of cumulative distribution function of re-projection error on the \textbf{SUN3D} dataset. \textit{RegNet} achieves the best alignment performance by aligning most image pairs to a smaller reprojection error.}
		\label{fig:ReprojectionErrorCDF}
	\end{figure}	
	
		\begin{figure}[htbp]
		\begin{minipage}[b]{1.0\linewidth}
			\centering
			\includegraphics[width=8cm]{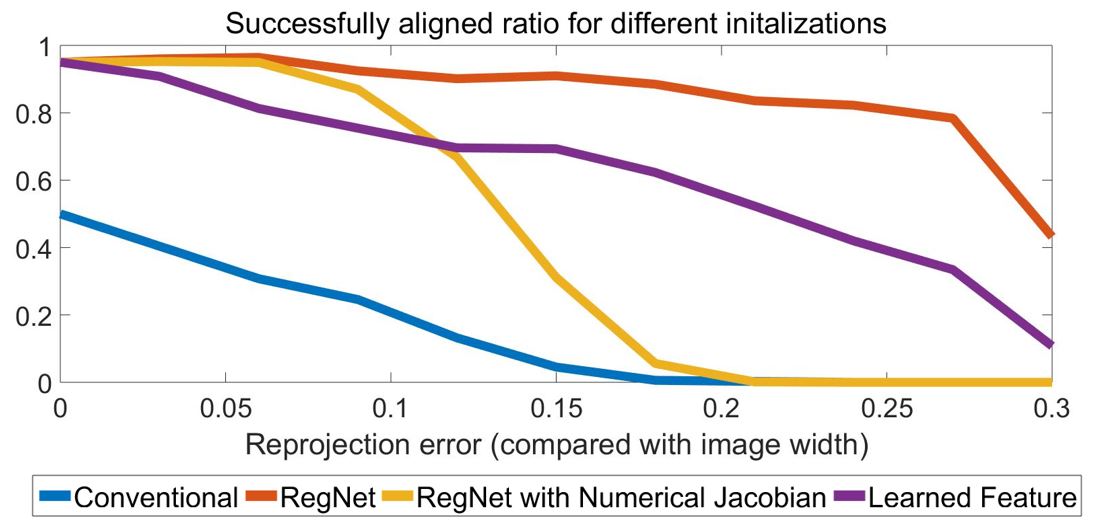}
		\end{minipage}
		\caption{Illustration of successfully registered image pair ratio versus different initial re-projection error. Image pairs with a smaller initial reprojection error are easier to be successfully registered. }
		\label{fig:SuccessfullyRatio}
	\end{figure}	


We use the test set of \textit{SUN3D} dataset to evaluate the alignment performance of different algorithms. 
Each test sample is tested 100 times with random distortions added in the initialization, and the optimization runs for 5 iterations. 
Image alignment quality is evaluated by the percentage of the re-projection error against the image width. 
The corresponding results are presented in Fig.~\ref{fig:ReprojectionErrorCDF} using the cumulative distribution function of the optimized re-projection error from different methods. 
The initial re-projection error without optimization is also presented as a reference. As shown in this experiment, 
\begin{itemize}
    \item \textit{Conventional} (using numerical Jacobian to minimize the photometric error) is only slightly better than the initial status, which proves that numerical Jacobian and photometric error is not applicable for large baseline image pair alignment problems.
    \item  \textit{Learned Feature} (using numerical Jacobian and learned feature) achieves comparable performance with \textit{RegNet} given a large threshold yet fails for more precise alignments, demonstrating that feature representation learned using numerical Jacobian is only able to roughly align image pairs. 
    \item \textit{RegNet} (joint learning of feature representation and Jacobian matrix) outperforms the others by inducing the smallest re-projection error. 
\end{itemize}
	\begin{figure*}[htbp]
		\begin{minipage}[b]{1.0\linewidth}
			\centering
			\includegraphics[width=18cm]{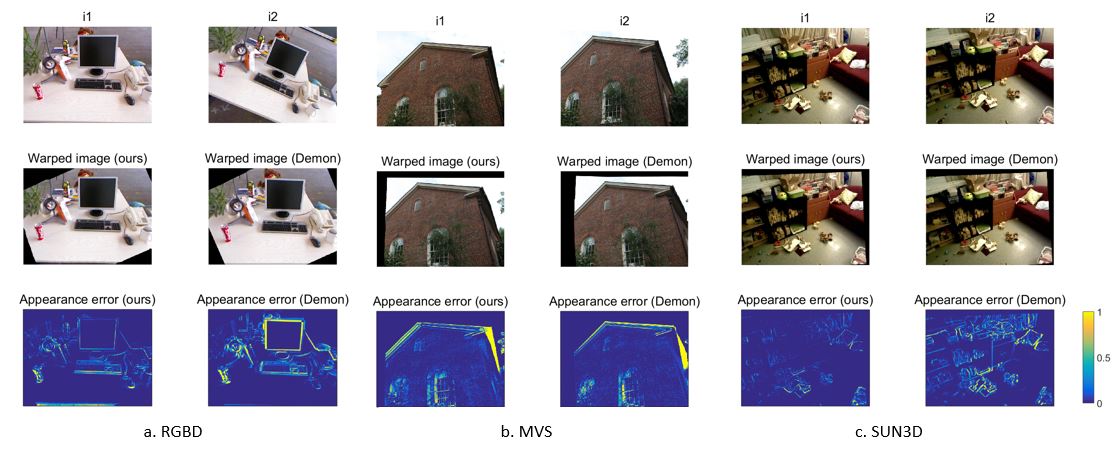}
		\end{minipage}
		\caption{An image pair is randomly chosen from the test set of each dataset and visualized here to show the image-to-image registration result compared with Demon. Image $i2$ is warped to the coordinate of $i1$ based on the predicted depth and motion parameters. Left is our result and right is the result from Demon. Appearance error is computed based on the photometric error between the warped image and the original image $i1$ (normalized to [0 1] for visualization).}
		\label{fig:RegExample}
	\end{figure*}

To further investigate the performance of the proposed Jacobian prediction network, we implement experiments to check the successfully registered image pair ratio versus different initial re-projection errors, as shown in Fig.~\ref{fig:SuccessfullyRatio}. We group image pairs based on their initial re-projection error for every $5\%$ of the image width, from $0$ to $0.3$, and the last group is set as $[0.3\quad1]$.
Image pairs whose re-projection error is below $5\%$ after optimization is regarded as a successful registration. It can be observed that
\begin{itemize}
    \item \textit{RegNet} achieves the best successful ratio for all the initialization conditions. The advantage of our RegNet tends to be more significant when the initial re-projection error is larger compared with the other algorithms. For extreme cases when the re-projection error is larger than $30\%$, most direct methods fail completely while joint learning of feature and Jacobian scheme achieves a $40\%$ successful rate. 
    \item \textit{Learned Feature} achieves the second place in average. More interestingly, compared with the \textit{RegNet with Numerical Jacobian} method, \textit{Learned Feature} achieves a higher successful rate when the initial reprojection error is large while a lower successful rate when the initial reprojection error is small. Such phenomena reveals that due to the forced smooth property as analyzed in Sec~\ref{sec_JPN}, feature representation learned in \textit{Learned Feature} is over-smoothed and less precise than the features learned using \textit{RegNet}. 
\end{itemize}

\subsection{Qualitative Evaluation}
We randomly select one test image pair in each dataset to compare the image-to-image registration result qualitatively. As shown in Fig.~\ref{fig:RegExample}, the column of a, b and c corresponds to the dataset of RGBD, MVS and SUN3D, respectively. For each dataset, the first row presents the input image pairs (left for $I_1$ and right for $I_2$). The second row shows the warped image from $I_2$ to the coordinate of $I_1$ based on our method (second row left) and Demon (second row right), respectively. Then the third row visualizes the registration error (represented by the photometric error between $I_1$ and the warped image of $I_2$). Apparently,
the proposed joint learning of feature and Jacobian scheme assures superior performance than Demon in producing lighter error map for image-to-image alignment.

\section{Conclusion}
In this paper, we presented RegNet, a novel neural network architecture that jointly learns Jacobian with the feature space, resulting in an increased convergence range for image-to-image alignment problem. Our ablation study demonstrates that jointly learning of feature and Jacobian improves the performance of direct image-to-image alignment optimization, especially for challenging cases with a poor initial alignment. For future works, we plan to extend the current image-to-image alignment to a full monocular SLAM system for dense monocular 3D reconstruction.
The proposed approach will be released publicly.

\section*{Acknowledgments}
This work was supported by the Natural Science Foundation of China under Grant 61722209 and Grant 61860206003. This work was supported by a Google Research Grant, a TUM Foundation Fellowship, a TUM-IAS Rudolf M{\"o}{\ss}bauer Fellowship, and the ERC Starting Grant \emph{Scan2CAD}.

{\small
\bibliographystyle{ieee}
\bibliography{egbib}
}

\end{document}